\documentclass[pdflatex,sn-mathphys-num,iicol]{sn-jnl}

\usepackage{graphicx}%
\usepackage{multirow}%
\usepackage{amsmath,amssymb,amsfonts}%
\usepackage{amsthm}%
\usepackage{mathrsfs}%
\usepackage[title]{appendix}%
\usepackage{xcolor}%
\usepackage{textcomp}%
\usepackage{manyfoot}%
\usepackage{booktabs}%
\usepackage{algorithm}%
\usepackage{algorithmicx}%
\usepackage{algpseudocode}%
\usepackage{listings}%
\usepackage{amsmath,amssymb}
\usepackage{algorithm}
\usepackage{algpseudocode}
\usepackage{booktabs} 
\usepackage{graphicx}

\theoremstyle{thmstyleone}%
\theoremstyle{thmstyletwo}%

\theoremstyle{thmstylethree}%

\theoremstyle{thmstyleone}%
\begin{document}

\title[Article Title]{Safe Reinforcement Learning of Autonomous Highway Driving: A Unified Framework for Safety and Efficiency}


\author[1]{\fnm{Chufei} \sur{Yan}}\email{yanchufei@nenu.edu.cn}

\author[2]{\fnm{Zhihao} \sur{Cui}}\email{2131523@tongji.edu.cn}

\author[1]{\fnm{Yiyan} \sur{Lv}}\email{lvyiyan123@nenu.edu.cn}

\author[1]{\fnm{Taojie} \sur{Chen}}\email{chentaojie@nenu.edu.cn}

\author[3]{\fnm{Ning} \sur{Bian}}\email{biann@dfmc.com.cn}

\author*[1]{\fnm{Yulei} \sur{Wang}\footnotemark}\email{wangyule@nenu.edu.cn}



\affil[1]{\orgdiv{School of Physics}, \orgname{Northeast Normal University},
\orgaddress{\city{Changchun}, \postcode{130024}, \country{China}}}

\affil[2]{\orgdiv{Clean Energy Automotive Engineering Center, School of Automotive Studies}, \orgname{Tongji University},
\orgaddress{\city{Shanghai}, \postcode{201804}, \country{China}}}

\affil[3]{\orgname{Mengshi Automobile Technology Company, Dongfeng Motor Corporation},
\orgaddress{\city{Wuhan}, \postcode{430010}, \country{China}}}


\abstract{Deep reinforcement learning (DRL) offers a compelling route to decision-making for advanced autonomous vehicles (AVs), yet its trial-and-error nature makes it difficult to guarantee safety during training and to achieve both safety and efficiency at deployment. We propose a unified safe reinforcement learning (SRL) framework that integrates safe distance (SD) \cite{7313361}, reward machines (RM) \cite{icarte2022reward}, and mixture-of-experts (MoE) \cite{shazeer2017outrageously}, termed MoE-RM-SRL. For deployment, SD and RM jointly shape a rule-aware reward that encodes highway traffic regulations and stage-wise objectives, enabling safe and reliable behavior without sacrificing efficiency. For training, we introduce a sparsely gated MoE layer comprising up to 11 deep $Q$-networks (DQNs); an SD-based gating rule activates a minimal set of experts for lane-keeping and lane-changing, mitigating the instability, discontinuities, and impulsive transients commonly induced by switching between heterogeneous controllers (e.g., MPC/rule-based modules and learned policies). We implement the proposed architecture in CARLA and integrate it with a 6-DoF driver-in-the-loop virtual-reality (DiL-VR) platform. Experiments in stochastic two-lane traffic show that MoE-RM-SRL substantially improves safety and efficiency over state-of-the-art baselines, and the framework naturally extends to multi-lane driving as well as on-ramp merging and exiting scenarios.}


\keywords{Autonomous Vehicle; Safe Reinforcement Learning; Highway Scenarios; Decision-Making; Safe Distance; Reward Machine; Mixture-of-Experts}



\maketitle
\footnotetext{Corresponding author: Yulei Wang. This work is supported by National Natural Science Foundation (NNSF) of China under Grant 62373281 and Shanghai Municipal Science and Technology Commission 23ZR1467700.}

\section{Introduction}\label{sec1}

Road traffic accidents remain one of the leading causes of injury and mortality worldwide. In highway environments, autonomous vehicles (AVs) must address a range of complex decision-making tasks, including lane keeping, lane changes, on-/off-ramp merging and diverging, and roundabout negotiation. Meanwhile, due to the uncertain behaviors of surrounding vehicles and the rarity of long-tail events, improving the safety and efficiency of decision-making in autonomous driving (AD) has become an increasingly important scientific and engineering challenge \cite{kiran2021deep, chen2024end}. 

Traditional decision-making pipelines typically rely on rule- or model-based architectures to generate high-level decisions and low-level control commands \cite{Kammel2008AnnieWAY,Ziegler2014Bertha,Zhang2019StateLatticeMPC}. By leveraging prior knowledge through detailed modeling and manual parameter tuning, these approaches can deliver strong performance within predefined operational design domains (ODDs). However, they often become brittle in complex traffic because hand-crafted heuristics scale poorly and may fail to generalize under uncertainty and unmodeled interactions \cite{peng2020path}. 

Deep reinforcement learning (DRL) offers an alternative paradigm for highway AD decision-making: instead of assuming an accurate \emph{a priori} system model, it learns data-driven policies through interaction with the environment. This paradigm has demonstrated strong self-learning capability in highly challenging driving tasks \cite{kiran2021deep, wurman2022gt}. Nevertheless, the intrinsic ``trial-and-error'' exploration of DRL remains fundamentally at odds with the safety-critical nature of AV deployment \cite{peng2022spil, lin2024fusion}.

Accordingly, safe reinforcement learning (SRL) has emerged as a key research direction. SRL methods explicitly incorporate safety constraints \cite{wachi2020safe} and/or shielding mechanisms \cite{9945670} during policy learning and execution to reduce unsafe exploration and improve reliability \cite{garcia2015comprehensive,10675394}. Depending on how safety mechanisms intervene in decision-making and learning, existing studies can be broadly grouped into two paradigms \cite{xu2022cpq}. 

\textbf{Soft-constrained SRL.} The first paradigm formulates SRL as a constrained optimization problem with \emph{soft} constraints, and solves it via Lagrange-multiplier networks or augmented Lagrangian methods \cite{Ma2025,Gao2025,zhang2021safe,Zhao2025}. By integrating constraints into policy updates or value learning, these approaches seek a practical trade-off between safety and efficiency. In many implementations, however, constraints are enforced through penalty terms or expectation constraints, which can still permit violations—especially during early training or under distribution shift \cite{Ma2025,Gao2025}. Beyond Lagrangian formulations, Zhang et al.~\cite{zhang2021safe} introduced Lyapunov-type policy constraints for safe motion planning, providing theoretical guarantees that state trajectories remain within a designated safe set. To further improve inference and generalization in dynamic traffic, Zhao et al.~\cite{Zhao2025} leveraged large language models (LLMs) to partition trust regions and constraint regions during policy updates. Nevertheless, LLM-based reasoning is only indirectly grounded in physical state and may hallucinate constraints or affordances, particularly under rare events or adversarial behaviors, thereby weakening safety assurances.

\textbf{Hard-constrained SRL.} The 2nd paradigm enforces \emph{hard} constraints to guarantee safety by construction \cite{mo2021safe,mirchevska2018high,9945670,Abd2025}. Mo et al.~\cite{mo2021safe} proposed a switching strategy that combines an RL agent with Monte Carlo tree search (MCTS): when predicted future states become risky, an MCTS-based planner is activated to guide safer exploration. Yang et al.~\cite{Yang2025} developed a safe and efficient self-evolving framework that couples model predictive control (MPC) with RL, where MPC provides safety-critical control guarantees while RL improves efficiency. Mirchevska et al.~\cite{mirchevska2018high} proposed an SD-based RL lane-change policy combined with formal safety verification, ensuring that only safe actions are selected at any time. In \cite{kalweit2020deep}, a constrained $Q$-learning formulation was further introduced to embed hard constraints directly into the $Q$-update, yielding stronger performance than reward shaping and Lagrangian optimization in their evaluated settings.

Despite this progress, many existing SRL-based AD decision-making systems remain best suited to relatively structured and well-defined traffic scenarios, and their self-learning and scalability are often insufficient for complex highway driving—such as coupled lane keeping and lane changing in dense multi-lane traffic, on-/off-ramp merging and exiting, and roundabout interactions.

For autonomous highway driving, SRL crucially depends on a principled and expressive reward design. In practice, automotive engineers often craft reward functions in a case-by-case manner \cite{mirchevska2018high,kalweit2020deep}, while the learning agent has no explicit access to the underlying task structure or the high-level intent encoded by the designer. To bridge this gap, we adopt a unified reward-specification framework based on reward machines (RMs) \cite{icarte2018rewardmachines,cam-etal-ijcai19,icarte2022reward,li2022noisy,li2024reward}. 

An RM is a finite-state machine that formalizes task progression and reward emission for an RL agent \cite{icarte2018rewardmachines}. By exposing temporal structure and enabling task decomposition, RMs can substantially simplify credit assignment and accelerate learning. Prior work has extended RMs beyond user-specified automata: Camacho et al.~\cite{cam-etal-ijcai19} showed how RM structure can be inferred from experience and exploited for partially observable RL, while Icarte et al.~\cite{icarte2022reward} developed methodologies for leveraging RM structure to improve RL efficiency and robustness. More recently, Li et al.~\cite{li2022noisy,li2024reward} investigated RM-based DRL under noisy and uncertain environments, further supporting the suitability of RMs for realistic driving settings.

Building on these advances, our previous work \cite{10397271} was, to the best of our knowledge, the first to develop an RM--DRL decision-making framework tailored to expressway scenarios. That RM--DRL formulation enabled explicit specification of temporally extended objectives—e.g., maintaining high speed while avoiding unsafe outcomes—across stochastic lane-keeping and lane-change maneuvers. In contrast to \cite{10397271}, the primary objective of the present work is to extend RM-DRL to \emph{safe} reinforcement learning by introducing mapping functions that integrate safety constraints into both learning and execution. The overarching motivation is to achieve near-optimal driving efficiency while rigorously respecting safety constraints throughout the entire training and deployment lifecycle.

In the second part of this study, we develop SRL algorithms for multiple highway scenarios using a Mixture-of-Experts (MoE) architecture. MoE has recently become a core design pattern for scaling learning systems—most prominently in large language models for language modeling, machine translation, and deep reasoning—and is increasingly being adopted in AD decision-making \cite{shazeer2017outrageously,guo2025deepseek,11142664}. 

\begin{figure*}[t]
  \centering
  \includegraphics[width=1\linewidth]{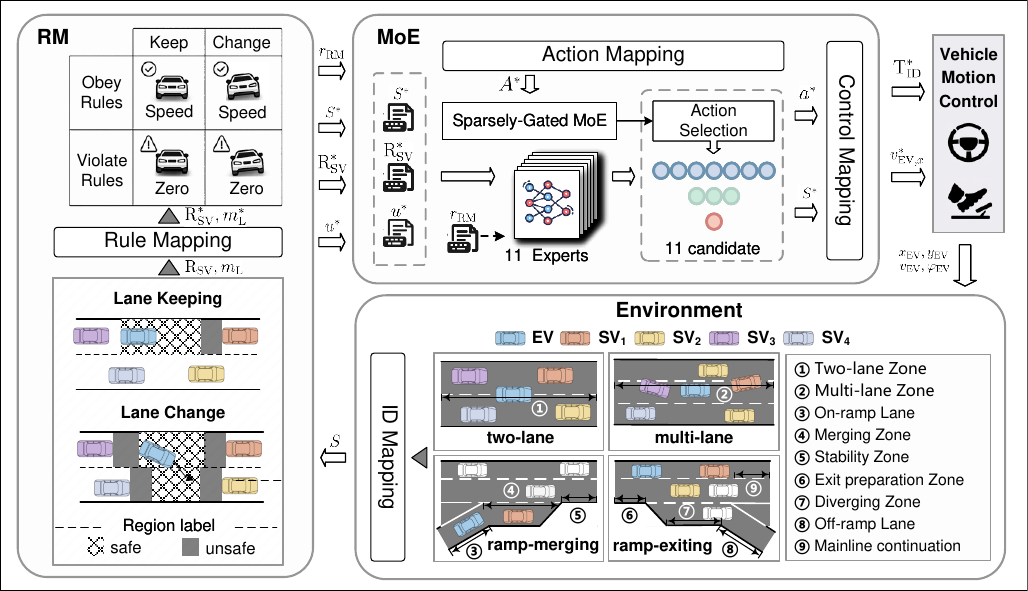}
  \caption{Overall architecture of the proposed MOE-RM-SRL framework}
  \label{fig:overview}
\end{figure*}

In this paper, we propose \textbf{MoE-RM-SRL}, a new SRL framework for autonomous highway decision-making (Fig.~\ref{fig:overview}). We first introduce the building blocks (safe distance (SD), RMs and mapping functions) and then unify them into a single architecture comprising up to 11 deep $Q$-networks (DQNs) coordinated by a sparsely gated MoE layer. Our exploration of MoE-centric SRL for highway driving was initiated in our preliminary study \cite{yan2025safe}; here, we substantially extend this line of work into a unified framework that supports multiple highway tasks, including two-lane driving, multi-lane driving, and on-ramp merging/exiting. 

We emphasize that our goal is to construct a safe and efficient self-evolving decision-making algorithm that learns \emph{directly} from recorded interaction data, rather than relying on extensive hand-engineered rules and scenario-specific tuning. In practical deployments, this design has the potential to reduce engineering overhead while improving safety and performance across diverse highway traffic conditions.

The remainder of this paper is organized as follows. Section~\ref{sec2} reviews the preliminaries, including scenario specifications, $Q$-learning, DQN, and RMs. Section~\ref{sec3} presents the proposed MoE-RM-SRL architecture in a systematic manner. Section~\ref{sec4} evaluates the proposed method through experiments on a driver-in-the-loop virtual-reality (DiL-VR) platform integrated with the CARLA simulator. We consider two-lane, multi-lane, and on-ramp merging/exiting scenarios under diverse traffic conditions, including different traffic densities, surrounding-vehicle driving styles, and adversarial events. We compare against several baseline methods to benchmark the proposed MoE-RM-SRL against the state of the art, and conduct ablation studies to quantify the contribution of each module. Finally, Section~\ref{sec5} concludes the paper and outlines directions for future work.

\section{Preliminaries}\label{sec2}

\subsection{Scenario Specifications}

This paper focuses on discrete decision-making for AVs in highway environments. We consider four representative scenarios: two-lane driving, multi-lane driving, on-ramp merging, and on-ramp exiting. The two-lane scenario is constructed in CARLA \texttt{Town04}, whereas the multi-lane and ramp scenarios are created in RoadRunner and imported into CARLA. Each scenario includes the ego vehicle (EV) and at least four surrounding vehicles (SVs). Low-level vehicle actuation and trajectory tracking of the EV are handled by CARLA’s default vehicle control settings.

To support structured reasoning and efficient spatial queries, we transform vehicle kinematics from the Cartesian frame to a Frenet reference frame and build the corresponding topological representation using a K-D tree. Note that the states of the EV and SVs are \emph{not} obtained via perceptual sensing. Instead, they are retrieved as ground-truth by querying vehicle actors directly from the CARLA world. The only sensor employed is a collision sensor (for episode termination and penalty assignment), rather than radar or LiDAR for detecting nearby vehicles.

\subsection{$Q$-learning}

We consider a Markov decision process (MDP) defined by the tuple
$\langle \mathcal{S}, \mathcal{A}, P, r, \gamma \rangle$,
where $\mathcal{S}$ and $\mathcal{A}$ denote the state and action spaces, respectively.
The transition kernel $P$ specifies the environment dynamics: when the agent is at state $s_t\in\mathcal{S}$ and executes action $a_t\in\mathcal{A}$ at time step $t$, the next state $s_{t+1}$ is sampled according to $P(\cdot | s_t, a_t)$.
The reward function $r$ provides a scalar feedback (e.g., $r_t=r(s_t,a_t,s_{t+1})$), and the discount factor $\gamma\in(0,1]$ weights future rewards relative to immediate rewards.

The MDP satisfies the Markov property: for any $t$, the distribution of the next state depends only on the current state-action pair,
\[
P(s_{t+1} | s_t,a_t,s_{0:t-1},a_{0:t-1}) = P(s_{t+1} | s_t,a_t).
\]
A (stochastic) policy $\pi(a | s)$ specifies a distribution over actions conditioned on the current state.
The agent seeks a policy that maximizes the expected discounted return
\[
J(\pi)=\mathbb{E}_{\pi}\!\left[\sum_{t=0}^{\infty}\gamma^t r_t\right].
\]

The action-value function $Q(s,a)$ estimates the expected return obtained by taking action $a$ in state $s$ and then following the policy thereafter. Greedy action selection is performed via
\begin{equation}\label{eq:greedy_action}
a = \arg\max_{a\in\mathcal{A}} Q(s, a).
\end{equation}
The optimal action-value function $Q^*(s,a)$ satisfies the Bellman optimality equation
\begin{align}
Q^*(s,a)= &~ \mathbb{E}_{s'\sim P(\cdot | s,a)}\Bigl[ r(s,a,s') \notag \\
&~~~ +\gamma\max_{a'\in\mathcal A}Q^*(s',a') \Bigr]. \label{eq:q_star}
\end{align}
$Q$-learning estimates $Q^*$ by iteratively minimizing the temporal-difference (TD) error using the TD target
\begin{equation}\label{eq:td_target}
y_t=r_t+\gamma\max_{a'\in\mathcal A}Q(s_{t+1},a'),
\end{equation}
and applying the update (with learning rate $\alpha\in(0,1]$)
\[
Q(s_t,a_t) \leftarrow (1-\alpha)\,Q(s_t,a_t) + \alpha\,y_t.
\]

In classical tabular $Q$-learning, $Q(s,a)$ is stored explicitly for each state-action pair. While effective for small discrete state spaces, this representation becomes impractical for continuous or high-dimensional states, motivating function approximation methods such as DQNs.

\subsection{Deep $Q$-network (DQN)}

Deep $Q$-networks (DQNs) extend $Q$-learning to continuous and high-dimensional state spaces by using a neural function approximator. Specifically, DQN employs a feedforward network $Q_{\text{net}}(s,a | \theta)$ to approximate the optimal action-value function $Q^*(s,a)$, where $s$ and $a$ denote the (vectorized) state and action representations, respectively. To improve training stability, DQN introduces (i) an experience replay buffer $\mathcal D$ that stores transition tuples and enables off-policy updates with decorrelated samples, and (ii) a target network $Q_{\text{net,target}}(s,a | \theta_{\text{target}})$ whose parameters are updated more slowly than those of the online network \cite{mnih2015human}.

Following the Bellman optimality equation, the temporal-difference (TD) target is constructed as
\begin{equation}\label{eq:dqn_target}
y_t=r_t+\gamma\max_{a'\in\mathcal A}Q_{\text{net,target}}(s_{t+1},a' | \theta_{\text{target}}),
\end{equation}
where $\theta_{\text{target}}$ denotes the parameters of the target network. The online network parameters $\theta$ are optimized by minimizing the squared TD error over transitions sampled from the replay buffer:
\begin{equation}\label{eq:dqn_loss}
\mathcal L(\theta)=
\mathbb E_{(s,a,r,s')\sim\mathcal D}
\Bigl[\bigl(y_t-Q_{\text{net}}(s,a | \theta)\bigr)^2\Bigr].
\end{equation}
During decision-making, DQN commonly adopts an $\epsilon$-greedy strategy.

\subsection{Reward Machine (RM)}

A reward machine (RM) is a finite-state-machine formalism for reward modeling. By providing an abstract, automaton-level description of task progress, an RM specifies how rewards are produced as a function of its internal state transitions. By binding reward emission to RM transitions, RMs can naturally represent temporally extended tasks and other non-Markovian specifications that depend on the history of interaction (i.e., task progress), even when the underlying environment dynamics are Markovian.

Formally, given a set of propositional symbols $\mathcal{P}$, an environment state space $\mathcal{S}$, and an action space $\mathcal{A}$, an RM is defined as a tuple
$\mathcal{R}_{\mathcal{P}\mathcal{S}\mathcal{A}}=\langle U, u_0, \delta_u, \delta_r \rangle$,
where $U$ is a finite set of RM states, $u_0\in U$ is the initial RM state, $\delta_u:U\times 2^{\mathcal{P}}\rightarrow U$ is the RM transition function, and $\delta_r:U\times 2^{\mathcal{P}}\rightarrow \mathbb{R}$ (or, more generally, $\delta_r:U\times 2^{\mathcal{P}}\times \mathcal{S}\times \mathcal{A}\times \mathcal{S}\rightarrow \mathbb{R}$) is the reward-output function.

At each time step $t$, the RM receives as input a truth assignment $\sigma_t \subseteq \mathcal{P}$, where $\sigma_t$ denotes the set of propositions that hold under the current environment state $s_t$. The RM then updates its internal state according to
$u_{t+1}=\delta_u(u_t,\sigma_t)$,
and emits a reward associated with the transition $(u_t \rightarrow u_{t+1})$ via $\delta_r$.

By augmenting the environment with the RM state, RMs impose a structured reward decomposition over an enlarged state space, enabling concise specification of complex rules and stage-wise objectives. In this work, we leverage RMs to express fine-grained, rule-aware reward logic for highway autonomous driving tasks.

\section{The Proposed Approach}\label{sec3}

In this section, we present a novel SRL approach for decision-making of autonomous vehicles (AVs) in highway environments.

\subsection{States and Actions}

\noindent\textbf{States.}
We adopt a Frenet coordinate system whose origin is anchored at the start of the lane reference line. The longitudinal displacement is denoted by $x$, the lateral displacement by $y$, the longitudinal velocity by $v_x$, and the lateral velocity by $v_y$. The lane width is $w_{\text{L}}$, and the longitudinal and lateral speed limits are $v_{x,\max}$ and $v_{y,\max}$, respectively. The lane index is denoted by $\text{L}_{\text{ID}}$. The perception range of the ego vehicle (EV) is $d_{\max}$, and the lane currently occupied by the EV is denoted as $\mathrm{C_{ID}}\in\{0,1\}$.

As illustrated in Fig.~\ref{fig:overview}, we define the EV state and the surrounding-vehicle (SV) states as
\begin{align}
s_{\mathrm{EV}}  &= \{x_{\mathrm{EV}}, y_{\mathrm{EV}}, v_{\mathrm{EV},x}, v_{\mathrm{EV},y}, \varphi_{\mathrm{EV}}\},\\
s_{\mathrm{SV}_i} &= \{x_{\mathrm{SV}_i}, y_{\mathrm{SV}_i}, v_{\mathrm{SV}_i,x}, v_{\mathrm{SV}_i,y}, \varphi_{\mathrm{SV}_i}\},
\end{align}
where $i\in\{1,\ldots,4\}$, $\varphi$ denotes the heading angle, $\mathrm{SV}_1$ and $\mathrm{SV}_3$ are the front and rear vehicles in the current lane, and $\mathrm{SV}_2$ and $\mathrm{SV}_4$ are the front and rear vehicles in the adjacent lane.

If a particular SV is not observed within the EV perception range, we introduce a virtual vehicle with a default state
\begin{align}
\mathrm{SV}_{i,\mathrm{Def}} =&~
\Bigl\{ x_{\mathrm{EV}}+\eta_i d_{\max},~
y_{\mathrm{EV}}+\lambda_i(1-2\,\mathrm{C_{ID}}) w_\text{L}, \notag\\
&\qquad \frac{1+\eta_i}{2}\,v_{x,\max},~0,~0 \Bigr\},
\end{align}
where $\eta_{1,2}=1$, $\eta_{3,4}=-1$, $\lambda_{1,3}=0$, and $\lambda_{2,4}=1$.
For the two-lane setting in Fig.~\ref{fig:overview}, $\text{L}_{\text{ID}}\in\{0,1\}$ denotes the left and right lanes, respectively.

Finally, we concatenate the EV and SV states to form the vehicle-level state vector:
\begin{equation}
s_{\mathrm{vehicle}}=
\{s_{\mathrm{EV}}, s_{\mathrm{SV}_{1}}, s_{\mathrm{SV}_{2}}, s_{\mathrm{SV}_{3}}, s_{\mathrm{SV}_{4}}\} \in \mathbb{R}^{25}.
\end{equation}

\noindent\textbf{Actions.}
The DQN outputs a discrete action $a  \in \mathcal{A}$ selected from the set
\begin{align}
\mathcal{A}= \{\text{Faster},\text{Idle},\text{Slower},\text{LaneLeft},\text{LaneRight} \}.
\end{align}
Here, $\text{Faster}$ commands an increase in the EV longitudinal speed by $\Delta v_{\text{A}}$, i.e.,
$\bar{v}_{\mathrm{EV},x}= v_{\mathrm{EV},x} + \Delta v_{\text{A}}$; 
$\text{Idle}$ maintains the current speed, $\bar{v}_{\mathrm{EV},x}=v_{\mathrm{EV},x}$; 
and $\text{Slower}$ commands a decrease in speed by $\Delta v_{\text{D}}$, i.e.,
$\bar{v}_{\mathrm{EV},x}= v_{\mathrm{EV},x} - \Delta v_{\text{D}}$.
The lane-change actions $\text{LaneLeft}$ and $\text{LaneRight}$ request a lane change to the left and right adjacent lanes, respectively.

\noindent\textbf{Augmented states.}
Let the current EV action be $a\in\mathcal{A}$, and introduce a binary indicator $\mathrm{m_L}\in\{0,1\}$ to denote whether the EV is close to the boundary between two lanes. Specifically, when
$y_{\mathrm{EV}}\in[w_\text{L}-d_{\text{ML}},\,w_\text{L}+d_{\text{ML}}]$,
we set $\mathrm{m_L}=1$; otherwise, $\mathrm{m_L}=0$.
Let $\mathrm{T_{ID}}\in\{0,1\}$ denote the EV’s target lane index, defined as
\begin{equation}\label{Eq16}
\mathrm{T_{ID}}=
\begin{cases}
\mathrm{C_{ID}}-1, &
\begin{aligned}[t]
\text{if } & a=\mathrm{LaneLeft} \land \mathrm{C_{ID}}=1,
\end{aligned}
\\
\mathrm{C_{ID}}+1, &
\begin{aligned}[t]
\text{if } & a=\mathrm{LaneRight} \land \mathrm{C_{ID}}=0,
\end{aligned}
\\
\mathrm{C_{ID}}, & \text{otherwise.}
\end{cases}
\end{equation}
In addition, let $v^{*}_{\mathrm{EV},x}$ denote the desired longitudinal speed of the EV.
The DQN input state is then formed by augmenting the vehicle state with these task-relevant variables:
\begin{equation}
s=\{s_{\mathrm{vehicle}},~\mathrm{C_{ID}},~\mathrm{m_L},~\mathrm{T_{ID}},~v^{*}_{\mathrm{EV},x}\}\in \mathbb{R}^{29}.
\end{equation}

\subsection{RM Design}\label{subsec3.2}

\noindent{\bf{Safe Distance}}:
In this work, the safe distance (SD) is defined as the \emph{critical longitudinal spacing} in a two-vehicle car-following setting such that, if the leading vehicle brakes with its maximum deceleration, the following vehicle can still avoid a collision.
Consider the ego vehicle ($\text{EV}$) and its leading vehicle $\text{SV}_i$ (e.g., $i=1$). Let $x_0\in\mathbb{R}$ denote a vehicle’s longitudinal position at time $t_0$, and let $a$ denote its longitudinal acceleration. For $t\ge t_0$, the longitudinal position evolves as
\[
x(t)=x_0+v_x(t-t_0)+\frac{1}{2}a(t-t_0)^2 .
\]
We define the longitudinal gap between the $\text{EV}$ and $\text{SV}_i$ as
$d_{\text{EV},\text{SV}_i}(t)=x_{\text{SV}_i}(t)-x_{\text{EV}}(t)$.
A collision occurs if there exists $t\ge t_0$ such that $d_{\text{EV},\text{SV}_i}(t)\le 0$.

By incorporating the human reaction time $\delta t$ into the SD formulation, we define
\begin{align}
\Delta x_{\text{SV}_i,\delta t} &= v_{\text{SV}_i,x}\,\delta t-\frac{1}{2}a_{\text{SV}_i,\text{D},\max}\,\delta t^2,\\
\Delta x_{\text{EV},\delta t} &= v_{\text{EV},x}\,\delta t,\\
v_{\text{SV}_i,\text{Res}} &= v_{\text{SV}_i,x}-a_{\text{SV}_i,\text{D},\max}\,\delta t,
\end{align}
where $\Delta x_{\text{EV},\delta t}$ is the displacement of the $\text{EV}$ over $\delta t$, $\Delta x_{\text{SV}_i,\delta t}$ is the displacement of $\text{SV}_i$ when braking at its maximum deceleration over $\delta t$, $v_{\text{SV}_i,\text{Res}}$ is the residual speed of $\text{SV}_i$ after the reaction interval, and $a_{\text{SV}_i,\text{D},\max}$ denotes the maximum longitudinal braking deceleration.

We define the minimum stopping distance of the $\text{EV}$ under maximum braking as
\begin{equation}
d_{\text{EV,SD}}=\frac{v_{\text{EV},x}^{2}}{2\lvert a_{\text{EV},\text{D},\max}\rvert},
d_{\text{SV}_i,\text{SD}}=\frac{v_{\text{SV}_i,x}^{2}}{2\lvert a_{\text{SV}_i,\text{D},\max}\rvert}.
\end{equation}
Under this setting, the minimum safe distance (SD) required between the $\text{EV}$ and $\text{SV}_i$ depends on whether $\text{SV}_i$ is longitudinally ahead of or behind the $\text{EV}$, which is captured by the following conditions:
\begin{align}
\forall i\in\{1,2\}:~ &(\Delta x_{\text{SV}_i,\delta t}\le d_{\text{EV,SD}})  \notag\\
&\wedge \bigl(\lvert a_{\text{SV}_i,\text{D},\max}\rvert< \lvert a_{\text{EV},\text{D},\max}\rvert\bigr)  \notag\\
&\wedge (v_{\text{SV}_i,\text{Res},x}<v_{\text{EV},x}) \notag\\
&\wedge (t_{\text{EV,Stop}}<t^{*}_{\text{SV}_i,\text{Stop}}), \label{eq:cond_front}\\
\forall i\in\{3,4\}:~ &(d_{\text{EV,SD}}\le \Delta x_{\text{SV}_i,\delta t}) \notag\\
&\wedge \bigl(\lvert a_{\text{EV},\text{D},\max}\rvert<\lvert a_{\text{SV}_i,\text{D},\max}\rvert\bigr) \notag\\
&\wedge (v_{\text{EV},x}<v_{\text{SV}_i,\text{Res},x}) \notag\\
&\wedge (t_{\text{EV,Stop}}<t^{*}_{\text{SV}_i,\text{Stop}}), \label{eq:cond_back}
\end{align}
where
\begin{equation}
t_{\text{EV,Stop}}=\frac{v_{\text{EV},x}}{\lvert a_{\text{EV},\text{D},\max}\rvert},
t^{*}_{\text{SV}_i,\text{Stop}}=\frac{v^{*}_{\text{SV}_i,x}}{\lvert a_{\text{SV}_i,\text{D},\max}\rvert}
\end{equation}
denote the braking-to-stop time of the $\text{EV}$ and the stopping time of $\text{SV}_i$, respectively.
Eqs.~\eqref{eq:cond_front} and~\eqref{eq:cond_back} use the same notation and differ only in whether $\text{SV}_i$ is located ahead of or behind the $\text{EV}$ along the longitudinal direction.

Subsequently, following \cite{pek2017verifying}, the SD between the $\text{EV}$ and $\text{SV}_i$ is defined as follows.
\begin{align}
& \text{Front vehicles}~(i=1,2). \notag \\
& d_{\text{SV}_i,\text{SD}1} = \Delta x_{\text{EV},\delta t}+d_{\text{EV,SD}},\\
& d_{\text{SV}_i,\text{SD}2} = d_{\text{SV}_i,\text{SD}1}-\Delta x_{\text{SV}_i,\delta t},\\
& d_{\text{SV}_i,\text{SD}3} =
\begin{cases}
\mathcal{B}_F -\Delta x_{\text{SV}_i,\delta t}, & \text{if Eq.~\eqref{eq:cond_front} holds,}\\
\mathcal{C}_F -d_{\text{SV}_i,\text{SD}}, & \text{otherwise,}
\end{cases} \\
& \mathcal{B}_F = \Delta x_{\text{EV},\delta t}
-\frac{\bigl(v_{\text{SV}_i,\text{Res},x}-v_{\text{EV},x}\bigr)^2}
{2\bigl(\lvert a_{\text{EV},\text{D},\max}\rvert-\lvert a_{\text{SV}_i,\text{D},\max}\rvert\bigr)}, \\
& \mathcal{C}_F = d_{\text{EV,SD}}+\Delta x_{\text{EV},\delta t}. 
\end{align}
\begin{align}
& \text{Rear vehicles}~(i=3,4). \notag \\
& d_{\text{SV}_i,\text{SD}1} = \Delta x_{\text{SV}_i,\delta t}+d_{\text{SV}_i,\text{SD}},\\
& d_{\text{SV}_i,\text{SD}2} = d_{\text{SV}_i,\text{SD}1}-v_{\text{EV},x}\delta t-\frac{1}{2}\lvert a_{\text{EV},\text{D},\max}\rvert\delta t^2,\\
& d_{\text{SV}_i,\text{SD}3} =
\begin{cases}
\mathcal{B}_R-v_{\text{EV},x}\delta t, & \text{if Eq.~\eqref{eq:cond_back} holds,}\\
\mathcal{C}_R+v_{\text{SV}_i,x}\delta t, & \text{otherwise,}
\end{cases} \\
& \mathcal{B}_R =
\frac{\bigl(v_{\text{EV},x}-\lvert a_{\text{EV},\text{D},\max}\rvert\delta t-v_{\text{SV}_i,x}\bigr)^2}
{2\bigl(\lvert a_{\text{SV}_i,\text{D},\max}\rvert-\lvert a_{\text{EV},\text{D},\max}\rvert\bigr)},\\
& \mathcal{C}_R =
\frac{v_{\text{SV}_i,x}^{2}}{2\lvert a_{\text{SV}_i,\text{D},\max}\rvert}
-\frac{v_{\text{EV},x}^{2}}{2\lvert a_{\text{EV},\text{D},\max}\rvert}.
\end{align}

By introducing the longitudinal relative distance and relative speed, we derive an explicit rule for determining whether the $\text{EV}$ maintains a safe distance to $\text{SV}_i$.
For $i\in\{1,2\}$ (SV ahead), define $\Delta x_{\text{SV}_i}=x_{\text{SV}_i}-x_{\text{EV}}$ and $\Delta v_{\text{SV}_i}=v_{\text{SV}_i,x}-v_{\text{EV},x}$;
for $i\in\{3,4\}$ (SV behind), define $\Delta x_{\text{SV}_i}=x_{\text{EV}}-x_{\text{SV}_i}$ and $\Delta v_{\text{SV}_i}=v_{\text{EV},x}-v_{\text{SV}_i,x}$.
Based on these quantities, the following formalized traffic-rule indicator is given by
\begin{align}
\text{R}_{\text{SV}_i} &=
\begin{cases}
0, & \text{if } \mathcal{D}=1,\\
1, & \text{if } \mathcal{E}=1,\\
(\Delta x_{\text{SV}_i}>d_{\text{SV}_{i},\mathrm{SD}3}), & \text{otherwise,}
\end{cases}
\end{align}
where
\begin{align}
\mathcal{D} &= \Bigl(\Delta x_{\text{SV}_i}\le \Delta v_{\text{SV}_i}\delta t + \tfrac{1}{2}\lvert a_{\text{SV}_i,\text{D},\max}\rvert\delta t^{2}\Bigr)\  \notag \\
& ~~~~~~ \vee\ (\Delta x_{\text{SV}_i}\le 0),\\
\mathcal{E} &= (\Delta x_{\text{SV}_i}>d_{\text{SV}_{i},\mathrm{SD}1})\ \vee\ \Bigl((\delta t\le t_{\text{SV}_i,\mathrm{Stop}}) \notag \\
& ~~~~~~ \wedge (\Delta x_{\text{SV}_i}>d_{\text{SV}_{i},\mathrm{SD}2})\Bigr).
\end{align}

\noindent{\bf{Reward machine (RM).}}
We now define the RM used in our highway-driving tasks. At each time step, the RM takes as input an abstracted signal of the current driving situation and outputs a reward for the RL agent. By decomposing highway driving into two modes—lane keeping and lane changing—we further distinguish traffic conditions as \emph{safe} or \emph{unsafe} with respect to the safe distance (SD) rules defined above. The resulting RM consists of four discrete states.

When the EV is in lane-keeping mode (i.e., the target lane coincides with the current lane), the RM checks whether the EV maintains an SD to the front vehicle in the current lane. If the SD condition is satisfied, the RM transitions to state $u_1$; otherwise, it transitions to state $u_2$. When the EV executes a lane change, the RM simultaneously evaluates the SD conditions with respect to all four SVs. If all safety conditions are satisfied, the RM enters state $u_3$; if any SD condition is violated, it enters state $u_4$.

Based on this design, the RM states are defined as
\begin{align}
u_1 &= \neg \mathit{m_L} \wedge \left( d_{\text{VL}} <\frac{w_{\text{L}}}{2} \right) \wedge \text{R}_{\text{SV}_1} \label{EqU1}\\
u_2 &= \neg \mathit{m_L} \wedge \left( d_{\text{VL}} <\frac{w_{\text{L}}}{2} \right) \wedge \neg \text{R}_{\text{SV}_1} \label{EqU2}\\
u_3 &= \mathit{m_L} \vee \left( d_{\text{VL}}  \ge \frac{w_{\text{L}}}{2}  \right) \wedge \left(\bigwedge_{i=1}^{4} \text{R}_{\text{SV}_i}\right) \label{EqU3}\\
u_4 &= \mathit{m_L} \vee \left( d_{\text{VL}} \ge \frac{w_{\text{L}}}{2} \right) \wedge \left(\bigvee_{i=1}^{4} \neg \text{R}_{\text{SV}_i}\right) \label{EqU4}
\end{align}
where $ d_{\text{VL}} =| y_{\text{EV}}-\mathrm{T_{ID}} w_\text{L} |$ is used to determine whether the EV is aligned with the target lane (lane keeping) or has deviated sufficiently to be considered in lane-change mode.

The RM then assigns the DQN reward as
\begin{align}
r_{\text{RM}} &=
\begin{cases}
\dfrac{v_{\text{EV},x}}{v_{\text{EV},x}^*}, & \text{if } u=u_1~\text{or}~u_3,\\[4pt]
0, & \text{otherwise,}
\end{cases} \\
v_{\text{EV},x}^* &=
\begin{cases}
v_{\text{SV}_1,x}, & \text{if}~(\mathcal{F}=1) \wedge \left(\displaystyle\bigvee_{i=1}^{4}\neg \text{R}_{\text{SV}_i}\right), \\
v_{x,\max}, & \text{otherwise,}
\end{cases}
\end{align}
where $\mathcal{F}= \{ \Delta x_{\text{SV}_1} \le d_{\mathrm{ACC}} \}$ and $v_{\text{EV},x}^*$ is the desired longitudinal speed and $d_{\mathrm{ACC}}$ denotes the distance threshold for adaptive cruise control (ACC).

\subsection{Mapping Design}

Compared with the two-lane setting, extending the proposed RM--RL scheme to more complex highway scenarios is non-trivial. To enable scalable reuse of the RM in Subsection~\ref{subsec3.2}, we introduce four mapping functions: \emph{ID mapping}, \emph{rule mapping}, \emph{action mapping}, and \emph{control mapping}. Collectively, these mappings adapt the two-lane RM formulation to multi-lane driving and on-ramp merging/exiting tasks.

\noindent\textbf{ID mapping.}
In the two-lane setting, the left and right lanes are indexed as $\text{L}_{\text{ID}}\in\{0,1\}$, respectively. For multi-lane highways (Fig.~\ref{fig:overview}), we decompose decision-making into at most two coupled two-lane subproblems. Specifically, when the $\text{EV}$ is in the leftmost or rightmost lane, there exists only one adjacent lane; we therefore treat that neighboring lane as the (single) adjacent lane and reduce the problem to a standard two-lane case. When the $\text{EV}$ is in an intermediate lane, we construct two two-lane models in parallel: one uses the left neighboring lane as the adjacent lane, and the other uses the right neighboring lane as the adjacent lane. At each decision step, the model that yields the larger $Q$ value is selected to output the decision.

For on-ramp merging (Fig.~\ref{fig:overview}), we partition the road along the longitudinal arc-length coordinate $x$ into an on-ramp lane $x\in[x_0,x_1)$, a merging zone $x\in[x_1,x_2)$, and a stability zone $x\in[x_2,x_3)$. When the $\text{EV}$ is on the on-ramp lane, we set its lane index to $\text{L}_{\text{ID}}=1$ and treat the mainline lane to its left as the adjacent lane with $\text{L}_{\text{ID}}=0$. Since there are no vehicles in the on-ramp's ``right-adjacent'' lane, we assign
$\text{SV}_2=\text{SV}_{2,\text{Def}}$ and $\text{SV}_4=\text{SV}_{4,\text{Def}}$.
When the $\text{EV}$ enters the merging zone, non-merging lane changes are prohibited; we therefore define the merging lane as $\text{L}_{\text{ID}}=1$ and its adjacent mainline lane as $\text{L}_{\text{ID}}=0$. Once the $\text{EV}$ reaches the stability zone, we revert to the standard two-lane indexing with the left lane as $\text{L}_{\text{ID}}=0$ and the right lane as $\text{L}_{\text{ID}}=1$.

For on-ramp exiting (Fig.~\ref{fig:overview}), we partition the road along the longitudinal arc-length coordinate $x$ into an exit-preparation zone $x\in[x_0,x_1)$, a diverging zone $x\in[x_1,x_2)$, a mainline continuation $x\in[x_2,x_{3,\mathrm{Fail}})$, and an off-ramp lane $x\in[x_2,x_{3,\mathrm{Suc}})$. The \emph{mainline continuation} denotes the subsequent mainline segment reached when the $\text{EV}$ fails to take the exit. When the $\text{EV}$ is in the diverging zone, we set $\text{L}_{\text{ID}}=0$ for the current lane and treat the right lane as the adjacent lane with $\text{L}_{\text{ID}}=1$, because non-exit lane changes are prohibited in this segment. When the $\text{EV}$ is on the off-ramp lane, we set its current lane as $\text{L}_{\text{ID}}=1$ and treat the left lane as the adjacent lane with $\text{L}_{\text{ID}}=0$; similarly, we assign $\text{SV}_2=\text{SV}_{2,\text{Def}}$ and $\text{SV}_4=\text{SV}_{4,\text{Def}}$ to reflect the absence of a right-adjacent lane. When the $\text{EV}$ is in either the exit-preparation zone or the mainline continuation, we use the standard two-lane indexing: left lane $\text{L}_{\text{ID}}=0$ and right lane $\text{L}_{\text{ID}}=1$.

\noindent\textbf{Rule mapping.}
We next introduce a modified criterion $m_{\text{L}}^{*}$ to determine whether a vehicle lies in the vicinity of the lane centerline region $y_{\text{lane}}=[-d_{\text{ML}},\,d_{\text{ML}}]$. Let the center position of vehicle $j$ be $\mathbf c_j=[x_j,\,y_j]^{\top}$ with heading angle $\varphi_j$, and define the planar rotation matrix
$\mathbf R(\varphi_j)=\begin{bmatrix}\cos\varphi_j & -\sin\varphi_j\\ \sin\varphi_j & \cos\varphi_j\end{bmatrix}$.
Assume a vehicle length $L$ and width $B$. The offsets of the four corner points in the vehicle frame are
\begin{equation}
\mathbf Q=
\begin{bmatrix}
\frac{L}{2} & \frac{L}{2} & -\frac{L}{2} & -\frac{L}{2} \\
\frac{B}{2} & -\frac{B}{2} & \frac{B}{2} & -\frac{B}{2}
\end{bmatrix}\in\mathbb{R}^{2\times 4}.
\end{equation}
Let $\mathbf 1=\begin{bmatrix}1&1&1&1\end{bmatrix}^{\top}\in\mathbb{R}^{4}$. Then the corner points of vehicle $j$ in the world frame are
\begin{equation}
\mathbf P_j=\mathbf c_j\mathbf 1^{\top}+\mathbf R(\varphi_j)\mathbf Q\in\mathbb{R}^{2\times 4},
\end{equation}
where the $k$-th corner coordinate is $\mathbf p_{j,k}=\mathbf P_j(:,k)$.
We define the distances from the corner points to the lane centerline region by
\begin{equation}
\delta_{j,k}=\bigl\lvert \mathbf e_2^{\top}\mathbf p_{j,k}-y_{\text{lane}}\bigr\rvert,~
\mathbf e_2=
\begin{bmatrix}
0 \\ 1
\end{bmatrix}.
\end{equation}
Accordingly, the minimum corner-point distance is
\begin{equation}
\delta_{j,\min}=\min_{k\in\{1,\ldots,4\}}\delta_{j,k}.
\end{equation}
The vehicle is considered to be near the two-lane centerline if
\begin{equation}
m_{\text{L}}^{*}
=(\delta_{j,\min}<\varepsilon_p)\ \vee\ (\lvert y_j-y_{\text{lane}}\rvert<\varepsilon_d), \label{Eq31}
\end{equation}
where $\varepsilon_p>0$ and $\varepsilon_d>0$ are thresholds for the corner-point distance and center-point distance, respectively.

Based on $m_{\text{L}}^{*}$ in \eqref{Eq31}, we define the rule mapping as
\begin{equation}\label{Eq32}
\text{R}^{*}_{\text{SV}_i}=
\begin{cases}
\text{R}'_{\text{SV}_1}\wedge \text{R}'_{\text{SV}_2}, & \text{if } m^{*}_{\text{L},\text{SV}_{i\in \{1,2\}}}=1,\\
\text{R}'_{\text{SV}_3}\wedge \text{R}'_{\text{SV}_4}, & \text{if } m^{*}_{\text{L},\text{SV}_{i\in \{3,4\}}}=1,\\
\text{R}'_{\text{SV}_i}, & \text{otherwise,}
\end{cases}
\end{equation}
where
\begin{equation}
\text{R}'_{\text{SV}_i}=
\begin{cases}
0, &
\begin{array}[t]{@{}l@{}}
\text{if }(m^{*}_{\text{L},\text{EV}}=1) \wedge (\lvert\Delta x_{\text{SV}_i}\rvert<\varepsilon_s) \\
\wedge~\neg\!\left(\text{R}_{\text{SV}_2} \wedge \text{R}_{\text{SV}_4}\right),~~ i\in\{2,4\},
\end{array}
\\[4pt]
\text{R}_{\text{SV}_i}, & \text{otherwise.}
\end{cases}
\end{equation}
where $\varepsilon_s>0$ is a longitudinal proximity threshold used to handle ambiguous configurations near the lane boundary.
Finally, by substituting $m^{*}_{\text{L}}$ in \eqref{Eq31} and $\text{R}^{*}_{\text{SV}_i}$ in \eqref{Eq32} into Eqs.~\eqref{EqU1}--\eqref{EqU4}, we obtain the modified RM states $u^{*}$, which extend RM--RL to more complex highway scenarios, including multi-lane driving and on-ramp merging/exiting.

\noindent\textbf{Action mapping.}
To reduce the output dimensionality of the DQN, we introduce a reduced action set
\begin{equation}
a^* \in \mathcal{A}^*=\{\text{Faster},~\text{Idle},~\text{Slower},~\text{LaneChange} \},
\end{equation}
where $\text{LaneChange}$ denotes an \emph{undirected} lane-change request, i.e., the policy commands a lane change without explicitly specifying left or right.

\noindent\textbf{Control mapping.}
Because the action space is reduced, the target-lane update in \eqref{Eq16} no longer applies. We therefore redesign the control mapping, whose core is the computation of the target lane index. When $a^*\in\{\text{Faster},\text{Idle},\text{Slower}\}$, the lateral plan remains unchanged and the target lane equals the current lane. When $a^*=\text{LaneChange}$, the target lane depends on the EV’s lane-position status and the safety indicators of nearby vehicles.

Specifically, the modified target lane index is defined as
\begin{align}
\mathrm{T}^{*}_{\text{ID}}=
\begin{cases}
\mathrm{C_{ID}}, & \text{if } a^*\in\{\text{Faster},\text{Idle},\text{Slower}\},\\[2pt]
1-\mathrm{C_{ID}}, &
\begin{array}[t]{@{}l@{}}
\text{if } (a^*=\text{LaneChange}) \\
~~~\wedge \mathit{m}^{*}_{\text{L},\text{EV}}=0,
\end{array}
\end{cases}  \label{Eq34}
\end{align}
and, when $a^*=\text{LaneChange} \wedge \mathit{m}^{*}_{\text{L},\text{EV}}=1$, we further refine $\mathrm{T}^{*}_{\text{ID}}$ as
\begin{align}
\mathrm{T}^{*}_{\text{ID}}&=
\begin{cases}
1-\mathrm{C_{ID}}, & \text{if } (\text{R}^{*}_{\text{SV}_3}=0 \wedge \text{R}^{*}_{\text{SV}_4}=1),\\
\mathrm{C_{ID}}, & \text{if } (\text{R}^{*}_{\text{SV}_3}=1 \wedge \text{R}^{*}_{\text{SV}_4}=0),\\
\mathrm{C_{ID}}, & \text{if }
\begin{aligned}[t]
&(\text{R}^{*}_{\text{SV}_3}=0 \wedge \text{R}^{*}_{\text{SV}_4}=0)\\
&\wedge (\Delta x_{\text{SV}_3}>\Delta x_{\text{SV}_4}),
\end{aligned}\\
1-\mathrm{C_{ID}}, & \text{if }
\begin{aligned}[t]
&(\text{R}^{*}_{\text{SV}_3}=0 \wedge \text{R}^{*}_{\text{SV}_4}=0)\\
&\wedge (\Delta x_{\text{SV}_3}<\Delta x_{\text{SV}_4}),
\end{aligned}\\
\mathrm{C_{ID}}, & \text{otherwise.}
\end{cases} \label{Eq35}
\end{align}
The modified target lane index in \eqref{Eq34}--\eqref{Eq35} can be passed directly to CARLA’s motion-control module to generate the corresponding reference trajectory. For brevity, we omit the implementation details of the low-level controller.

Based on the above mappings, the DQN input state is redefined as $s\in\mathcal{S}^*$, where
\begin{equation}
\mathcal{S}^{*}=\{s_{\mathrm{vehicle}},~\mathrm{C_{ID}},~m^{*}_\text{L},~\mathrm{T}^{*}_{\text{ID}},~v^{*}_{\text{EV},x}\}\in\mathbb{R}^{29}.
\end{equation}

\subsection{Sparsely-Gated MoE}

To achieve safe reinforcement learning (SRL), we further develop a sparsely-gated mixture-of-experts (MoE) architecture. The MoE comprises three layers of expert sub-networks. Within each layer, all sub-networks share the same action dimensionality, while across layers the action sets are progressively restricted. Overall, the MoE spans the collection of all non-trivial subsets of the reduced action set $A^{*}$. The core idea is to \emph{verify} candidate experts against the SD-based safety constraints in a layer-by-layer manner, and then select the first expert whose admissible action set is consistent with the current safety requirements to output the action.

\begin{figure}[h]
  \centering
  \includegraphics[width=\linewidth]{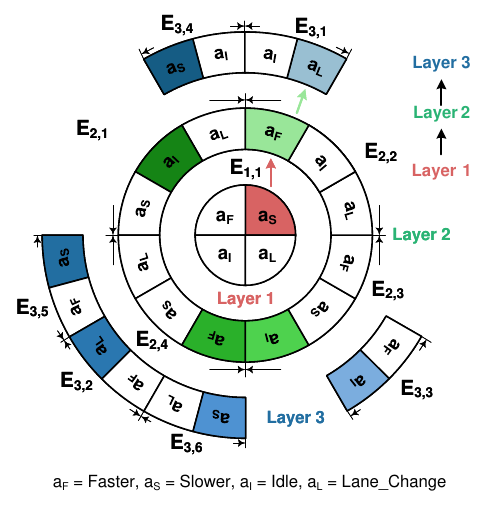}
  \caption{Schematic of the sparsely-gated MoE. The Layer-1 expert $\text{E}_{1,1}$ first outputs an unsafe action (rejected by Table~\ref{tab:safe_action_list}). The gating then activates the Layer-2 expert $\text{E}_{2,2}$, whose output is also rejected. Finally, the Layer-3 expert $\text{E}_{3,1}$ produces a safe action $a_{\text{L}}$, and the executed policy is $a_{\text{L}}=\text{Lane}\_\text{Change}$.}
  \label{fig:MOE}
\end{figure}

As illustrated in Fig.~\ref{fig:MOE}, the first layer contains a single expert $\text{E}_{1,1}$, which corresponds to the RM-RL DQN and outputs actions over the full reduced set:
\begin{equation}
A_{1,1}=A^{*}. \label{Eq37}
\end{equation}
The second layer consists of four experts $\text{E}_{2,1},\ldots,\text{E}_{2,4}$. Each expert inherits the same DQN backbone but operates on a 3-action subset obtained by removing exactly one action from $A^{*}$:
\begin{align}
A_{2,1} &= A^* \backslash \{\text{Faster}\}, \\
A_{2,2} &= A^* \backslash \{\text{Idle}\},\\
A_{2,3} &= A^* \backslash \{\text{Slower}\},\\
A_{2,4} &= A^* \backslash \{\text{LaneChange}\}.
\end{align}
Similarly, the third layer contains six experts $\text{E}_{3,1},\ldots,\text{E}_{3,6}$, each defined over a 2-action subset obtained by removing a distinct pair of actions from $A^{*}$:
\begin{align}
A_{3,1} &= A^* \backslash \{\text{Idle},\,\text{LaneChange}\}, \\
A_{3,2} &= A^* \backslash \{\text{Slower},\,\text{LaneChange}\},\\
A_{3,3} &= A^* \backslash \{\text{Slower},\,\text{Idle}\},\\
A_{3,4} &= A^* \backslash \{\text{Faster},\,\text{Idle}\},\\
A_{3,5} &= A^* \backslash \{\text{Faster},\,\text{LaneChange}\},\\
A_{3,6} &= A^* \backslash \{\text{Faster},\,\text{Slower}\}. \label{Eq47}
\end{align}

All 11 experts in \eqref{Eq37}--\eqref{Eq47} receive the same environment observation at each time step and compute their $Q$-values independently. Based on $\text{R}^{*}_{\text{SV}_i}$ in \eqref{Eq32}, we determine at time $t$ a safe action set $A_{\mathrm{safe}}\subseteq A^{*}$ by consulting the SD-based safety mask in Table~\ref{tab:safe_action_list}.

\begin{table}[h]
  \centering
  \caption{Safe actions in different conditions}
  \label{tab:safe_action_list}
  \scriptsize
  \setlength{\tabcolsep}{3.2pt}
  \renewcommand{\arraystretch}{1.05}
\begin{tabular}{c*{4}{c}*{4}{c}}
\toprule
\multirow[c]{2}{*}{No.} &
\multicolumn{4}{c}{Verification condition} &
\multicolumn{4}{c}{Action set} \\
\cmidrule(lr){2-5}\cmidrule(lr){6-9}
& $\text{R}^{*}_{\text{SV}_1}$ & $\text{R}^{*}_{\text{SV}_3}$ & $\text{R}^{*}_{\text{SV}_2}$ & $\text{R}^{*}_{\text{SV}_4}$
& $a_\text{F}$ & $a_\text{S}$ & $a_\text{I}$ & $a_\text{L}$ \\
\midrule
     1  & 1 & 1 & 1 & 1 & 1 & 1 & 1 & 1 \\
     2  & 1 & 1 & 1 & 0 & 1 & 1 & 1 & 0 \\
     3  & 1 & 1 & 0 & 1 & 1 & 1 & 1 & 0 \\
     4  & 1 & 1 & 0 & 0 & 1 & 1 & 1 & 0 \\
     5  & 1 & 0 & 1 & 1 & 1 & 0 & 1 & 1 \\
     6  & 1 & 0 & 1 & 0 & 1 & 0 & 1 & 0 \\
     7  & 1 & 0 & 0 & 1 & 1 & 0 & 1 & 0 \\
     8  & 1 & 0 & 0 & 0 & 1 & 0 & 1 & 0 \\
     9  & 0 & 1 & 1 & 1 & 0 & 1 & 0 & 1 \\
    10  & 0 & 1 & 1 & 0 & 0 & 1 & 0 & 0 \\
    11  & 0 & 1 & 0 & 1 & 0 & 1 & 0 & 0 \\
    12  & 0 & 1 & 0 & 0 & 0 & 1 & 0 & 0 \\
    13  & 0 & 0 & 1 & 1 & 0 & 0 & 0 & 1 \\
    14  & 0 & 0 & 1 & 0 & 0 & 0 & 1 & 0 \\
    15  & 0 & 0 & 0 & 1 & 0 & 0 & 1 & 0 \\
    16  & 0 & 0 & 0 & 0 & 0 & 0 & 1 & 0 \\
    \bottomrule
  \end{tabular}
{\footnotesize Note: 1 = safe, 0 = not safe, $a_{\text{F}}=\text{Faster}$, $a_{\text{S}}=\text{Slower}$, $a_{\text{I}}=\text{Idle}$, and $a_{\text{L}}=\text{LaneChange}$.} 
\end{table}

\begin{algorithm}[h!]
\caption{MOE-RM-SRL}
\label{alg:moe-rmrl}
\begin{algorithmic}[1]
\Require $\text{simulation environment}~\text{Env},\gamma \in [0,1]$
$\text{episode length}~\text{Step},$
$Q_{\text{tar}}~ \text{update interval}~n_t,$
steps of the $j$-th expert in the $i$-th group $n_{i,j}$,
CARLA low-level controller $C$.
environmental state $E_A=\langle S^*,\text{R}_{\text{SV}}^*,u^*\rangle$

\State Initialize replay buffer $\mathcal{D}_{i,j}$.
\State Initialize $Q_{i,j}(\cdot | \theta_{i,j})$,\: $Q_{\mathrm{tar},i,j}(\cdot | \theta_{\mathrm{tar},i,j})$.
\State Initialize $a^* \in A^* \Leftarrow \text{Action Mapping(}A\text{)}$.

  \For{$t=0,\ldots,\text{Step}-1$}
  \If{$t=0$}
    \State $s_{\text{vehicle}} \Leftarrow \text{ID Mapping(}\text{Env}\text{)}$.\ \label{ln:6}
    \State $\text{R}_{\text{SV}},~m_{\text{L}}^*,~\text{C}_\text{ID} \Leftarrow s_{\text{vehicle}}$ \label{ln:7}
    \State $u^*,\text{R}_{\text{SV}}^* \Leftarrow \text{Rule Mapping(}\text{R}_{\text{SV}},m_{\text{L}}^*\text{)}$.\label{ln:8}
    \State $S^* \Leftarrow (s_{\text{vehicle}},~\text{C}_\text{ID},~m_{\text{L}}^*,~\text{T}_\text{ID}^*~, v^{*}_{\text{EV},x})$\label{ln:9}
    \State $\text{T}_\text{ID}^*=\text{C}_\text{ID}$,~$v^{*}_{\text{EV},x}=0$.
    \State Initialize  $E_A$
  \Else
    \State $E_A = E_A'$.
  \EndIf
    \State $A_{\text{Safe}} \Leftarrow \text{R}_{\text{SV}}^* ,A_{\text{Safe}} \subseteq A^* $
    \State $a_{i,j}^* \Leftarrow \arg\max_{a^*\in A_{i,j}} Q_{i,j}(E_A, a^* | \theta_{i,j})$.
    \State $a_{\text{List}}^* = \{ a_{1,1}^*,...,a_{3,6}^*\}$.
    \State Sample $\xi\sim \mathrm{Uniform}(0,1)$.
    \If{$\xi<\varepsilon$}
    	\State $(a^*,i,j)\gets \mathrm{ExG}\bigl(A_{\text{Safe}},A_{i,j}\bigr)$.
    \Else
    	\State $(a^*,i,j)\gets \mathrm{PoG} (A_{\text{Safe}}, A_{i,j}, a_{\text{List}}^*)$.
    \EndIf
    \State $n_{i,j} = n_{i,j}+1$.
    \State $\text{T}_\text{ID}^*\Leftarrow \text{State Mapping(}a^*,S \text{)} $, $v^{*}_{\text{EV},x} \Leftarrow a^*$.
    \State $\text{Env},~ \text{Flag\_D} \Leftarrow C(\text{T}_\text{ID},v^{*}_{\text{EV},x})$.
    \State Compute the $E_A'$ at time $t+1$ by Lines \ref{ln:6}-\ref{ln:9}.
    \State Compute reward $r \Leftarrow r_{\text{RM}}(E_A, a^*, E_A')$.
    \State Substitute $(E_A,a^*,E_A',r)$ into $\mathcal{D}_{i,j}$.
    \State Sample $(E_A,a^*,E_A',r)$ from $\mathcal{D}_{i,j}$.
    \State $y=\begin{cases} r, & \text{If Flag\_D is ture},\\ r+ X, & \text{Otherwise}. \end{cases}$
    \State $X=\gamma\displaystyle\max_{a' \in A_{i,j}} Q_{\mathrm{tar},i,j} \bigr(E_A',a' | \theta_{\mathrm{tar},i,j}\bigr)$
   \If{$n_{i,j} > n_t$ and $j \neq 0$}
    \Statex \hspace*{3em} $Q_{\mathrm{tar},i,j}=Q_{i,j}, n_{i,j}=0$
    \EndIf
  \EndFor
\end{algorithmic}
\end{algorithm}

\begin{algorithm}[h]
\caption{ExG}
\label{alg:moe-explore}

\begin{algorithmic}[1]

      \State $\mathrm{act}_0 \gets \mathrm{Uniform}(A^*)$.
      \If{$\mathrm{act}_0 \in A_{\text{Safe}}$}
      	\State set $j\gets 1$
      	\State \Return $(\text{act}_{0},1,j)$
	\EndIf
	\State $j \leftarrow \{k\in\{1,\ldots,4\} | A_{2,k}\cap\{\mathrm{act}_0\}=\varnothing\}$
	\State $\mathrm{act}_{2,j} \gets \mathrm{Uniform}(A_{2,j})$
	\If{$\mathrm{act}_{2,j} \in A_{\text{Safe}}$}
      	\State \Return $(\text{act}_{2,j},2,j)$
	\EndIf
        \State $i \leftarrow \{k\in\{1,\ldots,6\} | A_{3,k}\cap\{\mathrm{act}_0,\mathrm{act}_{2,j}\}=\varnothing\}$
	\State $\mathrm{act}_{3,j} \gets \mathrm{Uniform}(A_{3,j})$
	\If{$\mathrm{act}_{3,j} \in A_{\text{Safe}}$}
      	\State \Return $(\text{act}_{3,j},3,j)$
	\ElsIf{$A_{3,j} \backslash \mathrm{act}_{3,j} \in A_{\text{Safe}}$}
		\State $\mathrm{act}_t  \leftarrow A_{3,j} \backslash \mathrm{act}_{3,j}$
		\State \Return $(\text{act}_{t},3,0)$
	\EndIf
	\State $\mathrm{act}_t \leftarrow \text{Idle}$, $i \leftarrow 0$
	\State \Return $(\text{act}_{t},3,0)$

\end{algorithmic}
\end{algorithm}

\begin{algorithm}[h]
\caption{PoG}
\label{alg:moe-polite}

\begin{algorithmic}[1]
      \If{$\mathrm{act}_{1,1} \in A_{\text{Safe}}$}
      	\State set $j\gets 1$
      	\State \Return $(\text{act}_{1,1},1,1)$
	\EndIf
	\State $j \leftarrow \{k\in\{1,\ldots,4\} | A_{2,k}\cap\{\mathrm{act}_{1,1}\}=\varnothing\}$
	\If{$\mathrm{act}_{2,j}\in A_{\text{Safe}}$}
      	\State \Return $(\text{act}_{i},2,j)$
	\EndIf
        \State $j \leftarrow \{ k \in \{1,\ldots, 6 \} | A_{3,k} \cap \{ \mathrm{act}_{1,1},\mathrm{act}_{2,j} \} = \varnothing \}$
	\If{$\mathrm{act}_{3,j} \in A_{\text{Safe}}$}
      	\State \Return $(\text{act}_{3,j},3,j)$
	\ElsIf{$A_{3,j} \backslash \mathrm{act}_{3,j} \in A_{\text{Safe}}$}
		\State $\mathrm{act}_t  \leftarrow A_{3,j} \backslash \mathrm{act}_{3,j}$
		\State \Return $(\text{act}_{t},3,0)$
	\EndIf
	\State $\mathrm{act}_t \leftarrow \text{Idle}$, $i \leftarrow 0$
	\State \Return $(\text{act}_{t},3,0)$
\end{algorithmic}
\end{algorithm}

At any time step, only one expert is executed. We therefore design the following sparsely-gated selection rule.

\begin{enumerate}
\item Query Table~\ref{tab:safe_action_list} to evaluate the action $a_{1,1}$ produced by $\text{E}_{1,1}$. If $a_{1,1}\in A_{\mathrm{safe}}$, activate $\text{E}_{1,1}$ and terminate the selection.
\item Otherwise, if $a_{1,1}\notin A_{\mathrm{safe}}$, evaluate the actions $a_{2,j}$ produced by the Layer-2 experts $\text{E}_{2,j}$ whose action sets exclude $a_{1,1}$, i.e., $a_{2,j}\in A^*\backslash\{a_{1,1}\}$. If there exists an expert such that $a_{2,j}\in A_{\mathrm{safe}}$, activate the corresponding $\text{E}_{2,j}$ and terminate the selection.
\item Otherwise, evaluate the actions $a_{3,k}$ produced by the Layer-3 experts $\text{E}_{3,k}$ whose action sets exclude both $a_{1,1}$ and the rejected Layer-2 action $a_{2,j}$, i.e., $a_{3,k}\in A^*\backslash\{a_{1,1},a_{2,j}\}$. If there exists an expert such that $a_{3,k}\in A_{\mathrm{safe}}$, activate $\text{E}_{3,k}$ and terminate the selection.
\item If none of the above actions is admissible, execute the remaining single action
$a\in A^*\backslash\{a_{1,1},a_{2,j},a_{3,k}\}$, which by construction satisfies $a\in A_{\mathrm{safe}}$, and terminate.
\end{enumerate}

Figure~\ref{fig:MOE} illustrates this sparsely-gated procedure through an example. The expert $\text{E}_{1,1}$ in Layer~1 first outputs an action that is deemed unsafe by Table~\ref{tab:safe_action_list}. The gating then activates $\text{E}_{2,2}$ in Layer~2, whose output is again rejected as unsafe. Subsequently, $\text{E}_{3,1}$ in Layer~3 outputs an admissible action $a_{\text{L}}$, and the final DQN policy becomes $a_{\text{L}}=\text{LaneChange}$.

Note that the proposed sparsely-gated MoE introduces neither handcrafted prior knowledge nor coupling between experts: all sub-networks are trained independently. Consequently, provided sufficient data, each expert inherits the standard convergence behavior of $Q$-learning, and the overall decision process remains well-defined and stable.

Finally, Algorithms~\ref{alg:moe-rmrl}--\ref{alg:moe-polite} summarize the proposed MoE-RM-SRL in pseudocode.

\begin{figure*}[t]
  \centering
  \includegraphics[width=1.0\linewidth]{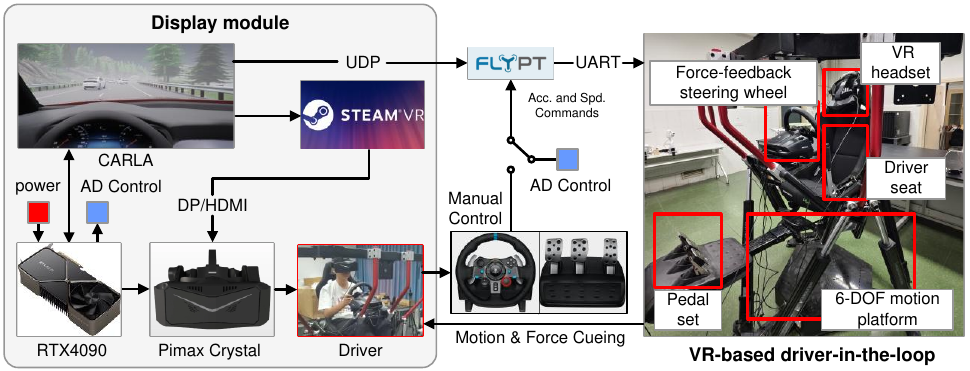}
  \caption{Schematic of the 6-DoF driver-in-the-loop VR simulator and experimental environment.}
  \label{fig:DiLVR}
\end{figure*}

\section{Experiments}\label{sec4}

\subsection{Scenario Description}

In this section, we validate the proposed SRL algorithm for highway decision-making and control on the 6-DoF driver-in-the-loop virtual-reality (DiL-VR) platform shown in Fig.~\ref{fig:DiLVR}. All training and evaluation are conducted in the CARLA simulator. The simulation maps include (i) a two-lane expressway segment in Town04, and (ii) a three-lane expressway segment as well as on-ramp merging/exiting segments constructed in RoadRunner and imported into CARLA. The lane width is fixed to $3.5~\text{m}$.

The ego vehicle (EV) is instantiated as a B-class passenger vehicle with full vehicle dynamics enabled. Traffic is generated in the vicinity of the EV, yielding a closed-loop traffic-flow setting centered on the agent. Unless otherwise specified, each episode is capped at $320$ simulation steps. Surrounding traffic is randomly initialized with speeds sampled from $[5,6]~\text{m}/\text{s}$. The EV speed is bounded by $v_{x,\max}=10~\text{m}/\text{s}$, and the perception range is set to $d_{\max}=50~\text{m}$. The lane-centerline tolerance for identifying the middle-lane region is $d_{\text{ML}}=1.75~\text{m}$. For the rule-mapping module, we set $\varepsilon_p=0.5~\text{m}$ and $\varepsilon_d=0.6~\text{m}$.

Algorithm hyperparameters are summarized in Table~\ref{tab:dqn_hparams}. All $Q$-networks share the same architecture, implemented as a three-layer fully connected neural network with 37 input neurons, two hidden layers of size $256\times 256$, and an output layer whose dimension is task-dependent (4/3/2 for the corresponding action subsets). Training and deployment are performed on a workstation equipped with an Intel Core U9-285K CPU and an NVIDIA GeForce RTX 4090 GPU.
\begin{table}[h!]
\centering
\caption{Parameters for the simulation}
\label{tab:dqn_hparams}
\renewcommand{\arraystretch}{1.15}
\begin{tabular}{@{} l l @{}}
\toprule
\textbf{Parameters} & \textbf{Symbol $\&$ Value} \\
\midrule
Simulation step size & $\delta t = 0.125~\text{s}$ \\
Learning rate & $1\times10^{-4}$ \\
Discount factor of network & $\gamma = 0.99$ \\
Vehicle length & $L=5~\text{m}$ \\
Vehicle width  & $B=2~\text{m}$ \\
Maximum acceleration & $a_{\text{EV},\text{A},\max}=3.5~\text{m}/\text{s}^2$ \\
Maximum deceleration & $a_{\text{EV},\text{D},\max}=6~\text{m}/\text{s}^2$ \\
Total training steps & $T_{\text{total}}=5 \times 10^5$ \\
Batch size & $n_b = 256$ \\
\bottomrule
\end{tabular}
\end{table}

\begin{table}[h!]
\centering
\caption{Parameters for driving styles}
\label{tab:driving_style}
\setlength{\tabcolsep}{10pt}
\begin{tabular}{lccc}
\toprule
 & Aggressive & Normal & Cautious \\
\midrule
DH & $L$ & $2L$ & $3L$ \\
SVR & \(30\%\) & \(-30\%\) & \(0\) \\
ALC & Yes & No & Yes \\
ABP & \( \frac{1}{9} \) & $0$ & \(\frac{1}{27}\)  \\
\bottomrule
\end{tabular}
{\footnotesize DH: Desired headway, SVR: Speed limit violation rate, ALC: Automated lane change, ABP: Adversarial behavior probability.}
\end{table}

\begin{table}[t]
\centering
\caption{Parameters for traffic density levels}
\label{tab:traffic_levels}
\setlength{\tabcolsep}{10pt}
\begin{tabular}{lccc}
\toprule
Traffic Level & Flow ($n_{\text{v}}$) & Headway (m)\\
\midrule
Level A & 8  & $[20,30]$\\
Level B & 16 & $[15,25]$\\
Level C & 24 & $[10,20]$\\
Level D & 32 & $[8,16]$\\
Level E & 40 & $[6,12]$\\
Level F & 44 & $[5,10]$\\
\bottomrule
\end{tabular}
\vspace{2mm}
\begin{minipage}{0.98\linewidth}
\footnotesize
\end{minipage}
\end{table}

\subsection{Metrics and Baselines}

We vary surrounding-vehicle (SV) driving styles and traffic densities, as summarized in Tables~\ref{tab:driving_style} and~\ref{tab:traffic_levels}. To enable a quantitative evaluation of the proposed method, we report four metrics that jointly characterize safety and efficiency during both deployment and training:
\begin{enumerate}
\item \textbf{Collision-free rate} $C_r$: the fraction of deployment episodes in which the EV completes an episode without any collision.
\item \textbf{Average speed}: the mean longitudinal speed of the EV over a deployment episode.
\item \textbf{Training collision-free rate} $T_c$: the fraction of training episodes in which the EV experiences no collision.
\item \textbf{Rise time of episode reward}: the number of training steps required for the episode return to reach $90\%$ of its peak value (used as a proxy for learning speed).
\end{enumerate}

To evaluate MOE-RM-SRL, we compare it against the several baselines across different traffic densities and simulation scenarios. The baselines are:
\begin{enumerate}
\item \textbf{Manual}: a human-driver baseline with 10 participants operating the DiL-VR platform under the same experimental conditions; we report the average performance across drivers.
\item \textbf{RL}: a standard DQN without reward machines; the reward is designed as a weighted combination of EV's speed and collision-related penalties.
\item \textbf{SR-RL}: the safe-action checking baseline in \cite{mirchevska2018high}, which iteratively tests actions in descending $Q$-value order and executes the highest-valued action that satisfies the safety condition.
\item \textbf{RM-RL}: the RM-based DQN baseline in \cite{10397271}, which combines reward machines with SD logic for highway driving.
\item \textbf{MP-RL}: an SR-RL variant augmented with the proposed mapping module (ID/rule/action/control mappings) to handle multi-lane and on-ramp scenarios.
\item \textbf{LM-RL}: a Lagrangian-DQN constrained-RL baseline \cite{achiam2017constrained}, widely used for safety-constrained policy learning and adopted in subsequent SafeRL benchmarks (e.g., PPO/TRPO variants) \cite{ray2019benchmarking}.
\item \textbf{ET-RL}: the expert-takeover baseline in \cite{wang2023adapt}, which activates a rule-based expert (implemented via the Intelligent Driver Model, IDM) to override unsafe behaviors.
\end{enumerate}

\subsection{Experimental Setup}

To evaluate the proposed method, we design three groups of experiments.

\noindent\textbf{Group 1: Two-lane with varying traffic densities.}
We consider a two-lane expressway segment under different traffic-density levels. The proposed method and all baselines are trained under a medium-density setting (Level~C) and then deployed and evaluated across all traffic levels. For each episode, the driving style of each SV is sampled uniformly at random from the three styles in Table~\ref{tab:driving_style}. Unless otherwise stated, all methods are evaluated over $100$ episodes for each test condition.

\noindent\textbf{Group 2: Multi-lane and on-ramp merging/exiting.}
This group assesses the scalability of the proposed algorithm to multi-lane driving and on-ramp scenarios. During both training and deployment, SVs are generated under traffic density Level~C, and their driving styles are sampled uniformly at random from Table~\ref{tab:driving_style}. Each method is evaluated over 10 independent runs, where each run consists of 100 episodes.

For the multi-lane scenario, following the ID-mapping strategy, decision-making is realized by combining two two-lane models: one for the left--middle lane pair and the other for the middle--right lane pair. When the EV is in the middle lane, we select the model that yields the higher $Q$ value at the current state.

For on-ramp merging, the reward is modified to explicitly account for task completion:
\begin{equation*}
r=
\begin{cases}
-320, & \text{if EV collides,}\\
0.7\,r_{\text{RM}} + 0.3\,r_{\text{merge}}, & \text{otherwise,}
\end{cases}
\label{eq:r_merge}
\end{equation*}
where $r_{\text{merge}}$ is a success indicator that becomes active when the EV merges from the on-ramp lane into the stability zone.

For on-ramp exiting, we similarly incorporate stage-wise completion rewards:
\begin{equation*}
r=
\begin{cases}
-320, & \text{if EV collides,}\\
0.5\,r_{\text{RM}} + \sum_{i=1}^{2} 0.25\,r_{\text{exit},i}, & \text{otherwise,}
\end{cases}
\label{eq:r_exit}
\end{equation*}
where $r_{\text{exit},1}$ and $r_{\text{exit},2}$ indicate successful entry into the adjacent lane and the off-ramp lane, respectively.

\noindent\textbf{Group 3: Adversarial environments and accelerated testing.}
To stress-test robustness under rare and hazardous interactions, we construct an adversarial driving environment. Specifically, we assign each SV in the adjacent lane an additional attribute $d_{\text{trig}}$. When the distance between the EV and $\text{SV}_i$ falls below $\lvert d_{\text{trig}}-d_{\text{SV}_i}\rvert$, the SV initiates a forced lane change with probability $\mathrm{ABP}$ (Table~\ref{tab:driving_style}). For each episode, we sample $\lvert d_{\text{trig}}-d_{\text{SV}_i}\rvert\in\{0,5,10\}$ uniformly at random for each SV. Each SV performs this probabilistic lane-change trigger at most once per episode.

\begin{figure}[h]
  \centering
  \includegraphics[width=1.0\linewidth]{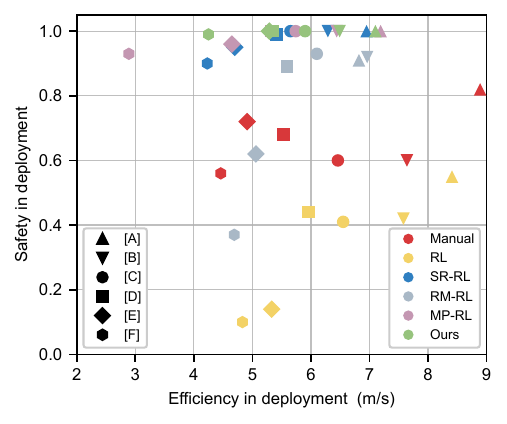}
  \caption{Comparison of the proposed algorithm and baselines in deployment.}
  \label{fig:ESinDP}
\end{figure}

\begin{figure}[h]
  \centering
  \includegraphics[width=1.0\linewidth]{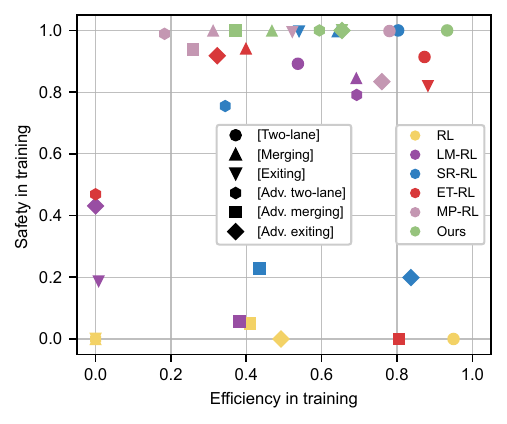}
  \caption{Comparison of the proposed algorithm and baselines in training.}
  \label{fig:ESinTN}
\end{figure}

Figures~\ref{fig:ESinDP} and~\ref{fig:ESinTN} summarize the comparative performance of the proposed method and baselines during deployment and training, respectively.

\subsection{Experimental Results}

Figure~\ref{fig:ESinDP} compares deployment performance of the evaluated RL algorithms across traffic-density levels A-F. We assess \emph{safety} using the collision-free rate and \emph{efficiency} using the average speed; the upper-right region therefore corresponds to better joint performance. Manual driving achieves moderate-to-high efficiency but exhibits non-uniform safety across density levels, highlighting the limitations of heuristic, threshold-based strategies as interaction complexity increases. The plain RL baseline attains high speeds, yet suffers the most pronounced degradation in safety as traffic becomes denser, indicating a brittle policy that does not transfer reliably to congested regimes. Safety-augmented baselines (SR-RL, MP-RL, and RM-RL) improve over RL but remain density-sensitive, with increased vertical dispersion under heavier traffic. This suggests residual constraint violations and/or conservative reactions that are not consistently effective across regimes. In contrast, MoE-RM-SRL consistently occupies the upper-right region across all traffic levels, indicating that it sustains near-ceiling deployment safety while preserving competitive driving efficiency, rather than trading speed for safety.

Figure~\ref{fig:ESinTN} continues to compare training-time behavior across several scenarios and conditions. The RL baseline lies almost entirely along the lower boundary (near-zero training safety) while spanning moderate-to-high training efficiency, reflecting extensive collision-prone exploration. LM-RL achieves high safety in some tasks but exhibits substantial variance, including near-collapse cases with very low safety and/or extremely low efficiency. In nominal tasks (two-lane, merging, and exiting), SR-RL clusters near high training safety with reasonable efficiency; however, under adversarial conditions (forced lane-change disturbances), SR-RL shows sharp drops in training safety, even when efficiency remains comparatively high. ET-RL attains high safety in several nominal settings but displays clear failure modes in at least one adversarial condition and does not consistently dominate the upper-right frontier. MP-RL remains largely within the high-safety band across tasks, albeit with moderate efficiency. Overall, the proposed MoE-RM-SRL provides the most favorable Pareto profile, broadly dominating the baselines by achieving both collision-free (or near collision-free) training and rapid reward rise, including under the forced-lane-change adversary.
\begin{table}[h]
\centering
\caption{Comparison of metrics in multi-lane scenarios}
\setlength{\tabcolsep}{12pt}
\label{tab:three_lane_results}
\begin{tabular}{lcc}
\toprule
\multirow{2}{*}{Metrics} & \multicolumn{2}{c}{Multi-lane} \\
\cmidrule(lr){2-3}
& Normal & Adversarial \\
\midrule
Ave. speed ($\text{m/s}$) & 7.34 & 6.69 \\
Collision rate ($\%$) & 0.00 & 13.4 \\
Ave. reward & 250.97 & 143.64 \\
\bottomrule
\end{tabular}
\end{table}
\begin{table*}[t]
\centering
\caption{Comprehensive comparison and evaluation of the proposed algorithm}
\label{tab:hard_results}
\setlength{\tabcolsep}{6pt}
\renewcommand{\arraystretch}{1.15}
\begin{tabular}{llcccccc}
\toprule
Scene & Metric & ET-RL & MP-RL & SR-RL & RL & LM-RL & Ours \\
\midrule
\multirow{4}{*}{$\text{Two-lane}$}
& $T_c$ & 0.914 & 0.998 & 1.000 & 0.000 & 0.895 & \textbf{1.000}\\
& $T_r$ & 97.31 & 228.64 & \textbf{229.10} & -165.38 & 162.18 & 228.19\\
& $C_r$ & 89.1\% & 100\% & 100\% & 8.6\% & 96.9\% & \textbf{100\%}\\
& $R_r$ & 90.42 & 214.00 & 213.07 & -161.38 & 168.51 & \textbf{217.89}\\
\midrule
\multirow{5}{*}{Merging}
& $T_c$ & 0.942 & 0.999 & 0.998 & 0.000 & 0.846 & \textbf{1.000}\\
& $T_r$ & 149.26 & 224.33 & 232.43 & -138.55 & 103.90 & \textbf{232.81}\\
& $C_r$ & 48.2\% & 100\% & 100\% & 97.3\% & 57.1\% & \textbf{100\%}\\
& $R_r$ & -65.04 & 207.74 & 238.45 & 182.49 & 20.49 & \textbf{240.42}\\
& $R_s$ & 89.3\% & 86.2\% & 87.0\% & 59.3\% & 57.1\% & \textbf{89.8\%}\\
\midrule
\multirow{5}{*}{Exiting}
& $T_c$ & 0.819 & 0.994 & 0.996 & 0.000 & 0.186 & \textbf{1.000}\\
& $T_r$ & 35.03 & 92.35 & 93.96 & -80.89 & -12.87 & \textbf{94.83}\\
& $C_r$ & 97.8\% & 99.4\% & 99.7\% & 62.1\% & 40.0\% & \textbf{100\%}\\
& $R_r$ & 35.70 & 58.90 & 51.42 & -39.44 & -78.08 & \textbf{72.52}\\
& $R_s$ & 0.10\% & 50.2\% & 42.3\% & 0.0\% & 0.0\% & \textbf{80.2\%}\\
\midrule
\multirow{4}{*}{$\text{Adv. two-lane}$}
& $T_c$ & 0.469 & 0.989 & 0.755 & 0.000 & 0.792 & \textbf{1.000}\\
& $T_r$ & -214.78 & 82.21 & -56.09 & -236.97 & -152.71 & \textbf{144.00}\\
& $C_r$ & 78.1\% & 87.5\% & 38.0\% & 5.0\% & 37.9\% & \textbf{92.9\%}\\
& $R_r$ & -57.41 & 100.23 & -72.05 & -238.13 & -68.97 & \textbf{155.03}\\
\midrule
\multirow{5}{*}{$\text{Adv. merging}$}
& $T_c$ & 0.000 & 0.938 & 0.364 & 0.047 & 0.057 & \textbf{1.000}\\
& $T_r$ & 72.68 & 85.30 & 90.00 & 47.50 & 14.17 & \textbf{95.39}\\
& $C_r$ & 85.3\% & 88.5\% & 89.7\% & 62.0\% & 48.7\% & \textbf{95.4\%}\\
& $R_r$ & 43.08 & 92.83 & 94.62 & 98.51 & 27.83 & \textbf{113.58}\\
& $R_s$ & 93.8\% & \textbf{97.1\%} & 95.6\% & 73.7\% & 81.7\% & 94.2\%\\
\midrule
\multirow{5}{*}{$\text{Adv. exiting}$}
& $T_c$ & 0.918 & 0.834 & 0.199 & 0.000 & 0.431 & \textbf{1.000}\\
& $T_r$ & -7.90 & 19.32 & -62.95 & -80.89 & -33.86 & \textbf{29.74}\\
& $C_r$ & 75.1\% & 85.8\% & 66.6\% & 27.5\% & 64.2\% & \textbf{92.7\%}\\
& $R_r$ & 6.44 & 21.18 & -0.09 & -91.18 & -36.82 & \textbf{35.40}\\
& $R_s$ & 10.1\% & 47.6\% & 53.4\% & 17.4\% & 0.1\% & \textbf{81.5\%}\\
\bottomrule
\end{tabular}
\end{table*}

Table~\ref{tab:three_lane_results} reports evaluation metrics for the proposed method in multi-lane scenarios. For the expressway setting, the average reward is computed as
\begin{equation*}
R_{r}=\sum_{i=1}^{n_{\max}} \left( \frac{1}{2} \frac{v_{\text{EV},x}}{v^*_{\text{EV},x}} +\frac{R^*_{\text{SV}_1}}{4} + \frac{R^*_{\text{SV}_2}}{4}  \right) -n_{\max} n_{\text{col}},
\end{equation*}
where $n_{\text{col}}$ indicates whether a collision occurs.
Under nominal multi-lane traffic, the proposed framework is both reliable and efficient, achieving collision-free operation at a comparatively high cruising speed. Under adversarial disturbances, the method retains a substantial fraction of its efficiency (only a single-digit percentage reduction in speed), but safety degrades materially, as evidenced by a non-trivial collision rate ($13.4\%$).

To provide a more comprehensive evaluation, we additionally report the following metrics: the number of training collisions $T_c$, the final training return $T_r$, the deployment collision rate $R_c$, the average deployment reward $R_r$, and the deployment success rate for merging/exiting $R_s$. For the on-ramp merging and exiting tasks, the deployment reward definitions are adjusted to reflect task completion, namely,
\begin{equation*}
R_{r}=\frac{7}{10} \sum_{i=1}^{n_{\max}} \frac{v_{\text{EV},x}}{v^*_{\text{EV},x}}
+\frac{3}{10} \sum_{i=1}^{n_{\max}} r_{\text{merge}}
-n_{\max}n_{\text{col}},
\end{equation*}
and
\begin{equation*}
R_{r}=\frac{1}{2} \sum_{i=1}^{n_{\max}} \frac{v_{\text{EV},x}}{v^*_{\text{EV},x}}
+\frac{3}{10} \sum_{i=1}^{n_{\max}} \sum_{j=1}^{2} r_{\text{exit},j}
-n_{\max}n_{\text{col}},
\end{equation*}
respectively.

Table~\ref{tab:hard_results} reports cross-scenario results for two-lane driving, on-ramp merging, on-ramp exiting, and their adversarial counterparts, covering training safety, training performance, and deployment outcomes. Overall, the results indicate that MoE-RM-SRL operates in a near Pareto-dominant regime: it achieves (near) collision-free training across almost all tasks while maintaining high deployment safety and state-of-the-art task efficiency. The most pronounced gains arise in the exiting and adversarial settings, where competing baselines typically sacrifice safety, task completion, or both.

\subsection{Ablation Study}

Finally, we perform an ablation study to quantify the contribution of each major component by removing the Mapping, MoE, and RM modules, respectively. All ablations are evaluated in the adversarial two-lane scenario. The results in Table~\ref{tab:ablation_S1_vertical_block} validate the necessity of each module and highlight their complementary roles in achieving safe and efficient learning and deployment.

\begin{table}[h!]
\centering
\caption{Ablation results (adv. two-lane)}
\label{tab:ablation_S1_vertical_block}
\setlength{\tabcolsep}{5pt}
\renewcommand{\arraystretch}{1.08}
\begin{tabular}{lcccc}
\toprule
\multirow{2}{*}{Variant} & \multicolumn{2}{c}{Training} & \multicolumn{2}{c}{Deployment} \\
\cmidrule(lr){2-3}\cmidrule(lr){4-5}
& $\text{T}_c $ & $\text{T}_r $
& $\text{C}_r $ & $\text{R}_r $ \\
\midrule
Full (Ours) & \textbf{1}  & \textbf{144}     & \textbf{92.9\%}   & \textbf{155.03} \\
w/o MAP     & 0           & -100.79          & 37.7\%            & -82.52 \\
w/o MoE     & 0.743       & 25.22            & 38.80\%           & 25.75 \\
w/o RM      & 0.983       & 164.00           & 87.1\%            & 70.03 \\
\bottomrule
\end{tabular}
\end{table}

\section{Conclusion and Future Work}\label{sec5}

This paper proposes a safe and efficient SRL algorithm for decision-making and control in AD systems. The proposed approach formulates decision-making as a unified design of RMs, MoE architecture, and mapping functions. This integration not only accelerates and stabilizes RL performance improvement, but also enforces safety throughout the learning process. The method is evaluated in two-lane and multi-lane highway settings, as well as on-ramp merging and exiting scenarios. Experimental results demonstrate that, compared with existing RL methods, the proposed algorithm achieves better performance while maintaining collision-free behavior during both training and deployment.

For future work, we plan to incorporate perception and planning modules and develop an end-to-end SRL-based driving framework to advance toward real-world experimentation. We will also further integrate domain knowledge with RL to enable more explainable performance gains, while supporting safe self-evolution without incurring prohibitively slow convergence.

\section*{Acknowledgment}
The authors would like to thank Prof. Hong Chen (0000-0002-1724-8649) and Prof. Yanjun Huang (0000-0003-3133-8031) of Tongji University for the earlier discussions of the paper idea.

\bibliography{refs}

\end{document}